\documentclass[lettersize,journal]{IEEEtran}

\IEEEoverridecommandlockouts   

\usepackage{cite}
\usepackage[pdftex]{graphicx}
\usepackage{amsmath}
\usepackage{dblfloatfix}
\usepackage{multirow}
\usepackage{booktabs}
\usepackage[flushleft]{threeparttable}

\usepackage{amsmath,amsfonts}
\usepackage{algorithmic}
\usepackage{algorithm}
\usepackage{array}
\usepackage[caption=false,font=normalsize,labelfont=sf,textfont=sf]{subfig}
\usepackage{textcomp}
\usepackage{url}
\usepackage{verbatim}
\usepackage{graphicx}
\usepackage{hyperref}
\def\BibTeX{{\rm B\kern-.05em{\sc i\kern-.025em b}\kern-.08em
    T\kern-.1667em\lower.7ex\hbox{E}\kern-.125emX}}
\usepackage{balance}

\usepackage{xcolor}
\newcommand{\hl}[1]{{#1}}

\begin{document}


\title{Leg Shaping and Event-Driven Control of a Small-Scale, Low-DoF Two-Mode Robot}

\author{Dingkun Guo,
        Larissa Wermers,
        and Kenn R. Oldham 

\thanks{Code and supplementary videos are available on the project website: \url{https://dkguo.com/research/walk2roll}.}
\thanks{This work was accepted by IEEE/ASME Transactions on Mechatronics on April 26, 2022. See details at: \url{https://doi.org/10.1109/TMECH.2022.3173183}}

\thanks{All authors are with the Department of Mechanical Engineering,
        University of Michigan, 2350 Hayward St, Ann Arbor, MI 48109
        (email: {\tt\small \{guodk, lwermers, oldham\}@umich.edu})}%

}

\maketitle

\begin{abstract}
Among small-scale mobile robots, multi-modal locomotion can help compensate for limited actuator capabilities.   However, supporting multiple locomotion modes or gaits in small terrestrial robots typically requires complex designs with low locomotion efficiency.   In this work, legged and rolling gaits are achieved by a 10~cm robot having just two degrees of freedom (DoF).  This is acheived by leg shaping that facilitates whole body rolling and event-driven control that maintains motion using simple inertial sensor measurements.  Speeds of approximately 0.4 and 2.2 body lengths per second are achieved in legged and rolling modes, respectively, with low cost of transport.   The proposed design approach and control techniques may aid in design of further miniaturized robots reliant on transducers with small range-of-motion.  
\end{abstract}


\section{Introduction}
 
 Miniaturizing mobile robots offers potential benefits for portability, cost, and access to confined spaces.   Many small-scale robots use bio-inspired designs and/or locomotion strategies.    However, expanding strategies for effective locomotion is desirable, as it is difficult to fully imitate many biological organisms within constraints of current actuation technology.  
 
Multi-modal locomotion offers opportunities to adapt to an environment even when individual modes have significant deficiencies.   Among terrestrial robots, various designs have been proposed for transitions between legged and rolling gaits, either through use of both wheels and legs, or by designing the entire robot for rolling motion (as through a cylindrical or spherical form).  Whole-body rolling examples include robots having legs that can be extended from within a spherical body~\cite{Aoki2020}, having legs that can be arranged to form a cylindrical or spherical body~\cite{Nemoto2015}\cite{Jia2017}\cite{Miura2019}, or having multi-link bodies that can approximate either circular or legged body plans~\cite{Phipps2008}.  These robots require many degrees-of-freedom (DoF) and tend to be complex; sizes range from about 20 to 70~cm.  

Reducing size and complexity of multi-modal robots is challenging with respect to both actuation and control.  As size decreases, wheeled locomotion becomes less desirable, due to difficulty of wheel assembly and disadvantageous scaling of electromagnetic transduction.   Transducers that do tend to scale more effectively, such as piezoelectric, shape memory alloy, and electrostatic devices, feature reasonable work capacities but low maximum strain, and hence finite range of motion.  Meanwhile, each additional DoF in a miniature robot is expensive in terms of space, sensing, and computational requirements, so use of few DoF is desirable.  Multi-modal locomotion has been demonstrated by a few robots having some of the above attributes.   Lin et al. described a robot modeled after a rolling caterpillar using a few addressable body segments~\cite{Lin2011}, and Duduta et al. described a robot based on a single dielectric elastomer capable of rolling, walking, and jumping~\cite{Duduta2020}.   However, these examples are operated in open-loop, with limited capacity for sustained or repeated movements. 
\begin{figure}[!t]
\centering
\includegraphics[width=\linewidth]{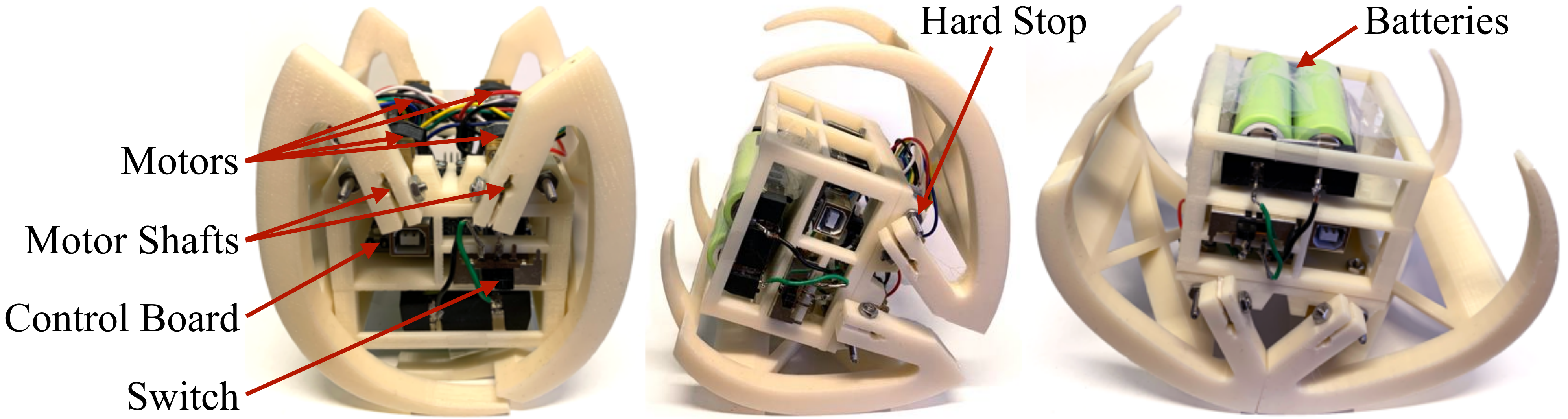}
\caption{Prototype 2-DoF walking-rolling robot.  Representative rolling phases: upright stance with legs closed (left); passive roll onto front leg (middle); legs open to roll onto rear leg (right).}
\label{fig:portrait}
\end{figure}
Here, we examine the design of curved leg geometries and an associated controller to produce efficient rolling by a small, terrestrial, two-mode robot.   The robot is developed as a test bed for locomotion by miniature robots based on curved appendages while limiting the degrees-of-freedom (DoF) and actuator range of motion required.  While the prototype robot is only modestly smaller than most prior works (11~cm at rest) and uses conventional DC gearmotors for actuation, its design is motivated by features that could facilitate further miniaturization.  The primary design novelty is the tailoring of leg curvature and an event-based controller based on a simple quasi-static analysis of passive gravitational contribution to rolling motion.  The resulting robot (Fig.\,\ref{fig:portrait}) is capable of legged and rolling locomotion with low cost-of-transport (CoT) relative to other small multi-modal robots. 

\section{Design}

This section introduces the robot concept and the procedure for design evaluation.  In full, robot design and gait optimization is a complex problem that could involve dynamic simulation in multiple modes and environments.   However, here we isolate the contribution of gravity to rolling locomotion in a preferential direction, to create simple metrics for leg geometries that facilitate whole-body rolling. 

The proposed robot architecture consists of two single-DoF legs, at the front and rear of the robot.  Through appropriate opening and closing, these can provide a continuous curved surface during robot rolling.  Alternatively, when upright, sequential motion of the front and rear legs can produce a simple walking gait.  This arrangement is inspired by generic curved, cantilever-like leg structures that can be fabricated from planar laminate materials at small dimensions through lithography processes.  To constrain the design space, we limit consideration to symmetric legs with a single rotary joint each. Asymmetric legs and legs that bend over distributed regions are potential variations that are not considered here.

Conceptually, rolling locomotion involves several steps.  Motion can be initiated from upright by unbalancing the CoM over closed legs.  Without further leg movement, the robot would then settle at an equilibrium position with the CoM at a low point over to the curved legs.  This can be overcome through a combination of momentum and actuator forcing.   Entering this study, we anticipated that rolling would be sustained primarily by ``kicking" backwards with the rear leg.  However, in the final design, motion is sustained largely by keeping the CoM ahead of the leg's contact with ground as much as possible, allowing a gravitational moment to add rotational momentum in the desired direction.  Resulting leg optimization focuses on manipulating CoM location through change in leg angle to enhance the gravitational moment in favorable positions (i.e., CoM ahead of the ground contact point) and reduce the gravitational moment in unfavorable positions (i.e., CoM behind the ground contact point.
\subsection{Parameterization and Evaluation Metrics}
\begin{figure}[!t]
\raggedright
\subfloat[]{\includegraphics[width=0.48\linewidth]{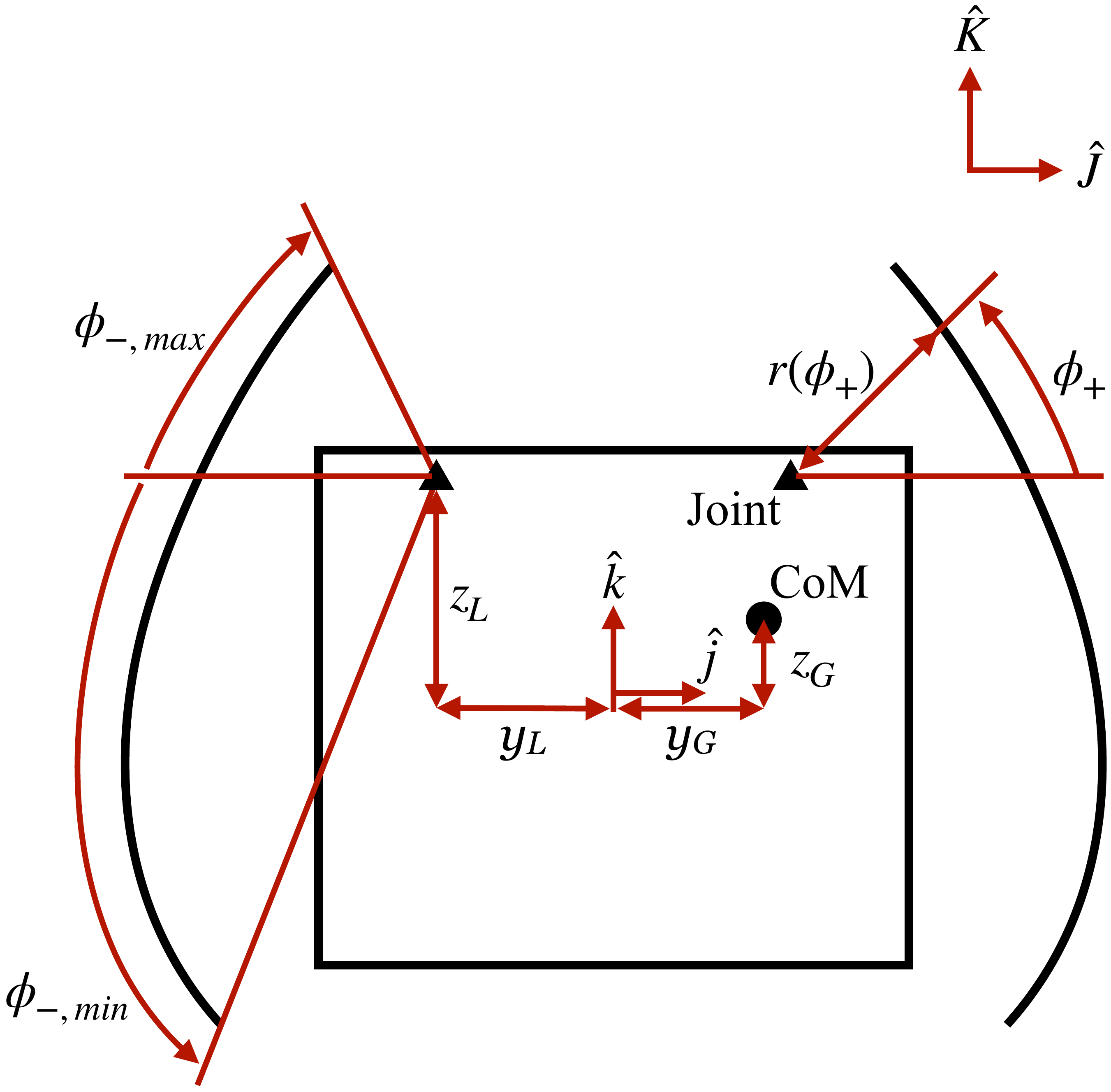}\label{fig:pa}}
\hfill
\subfloat[]{\includegraphics[width=0.50\linewidth]{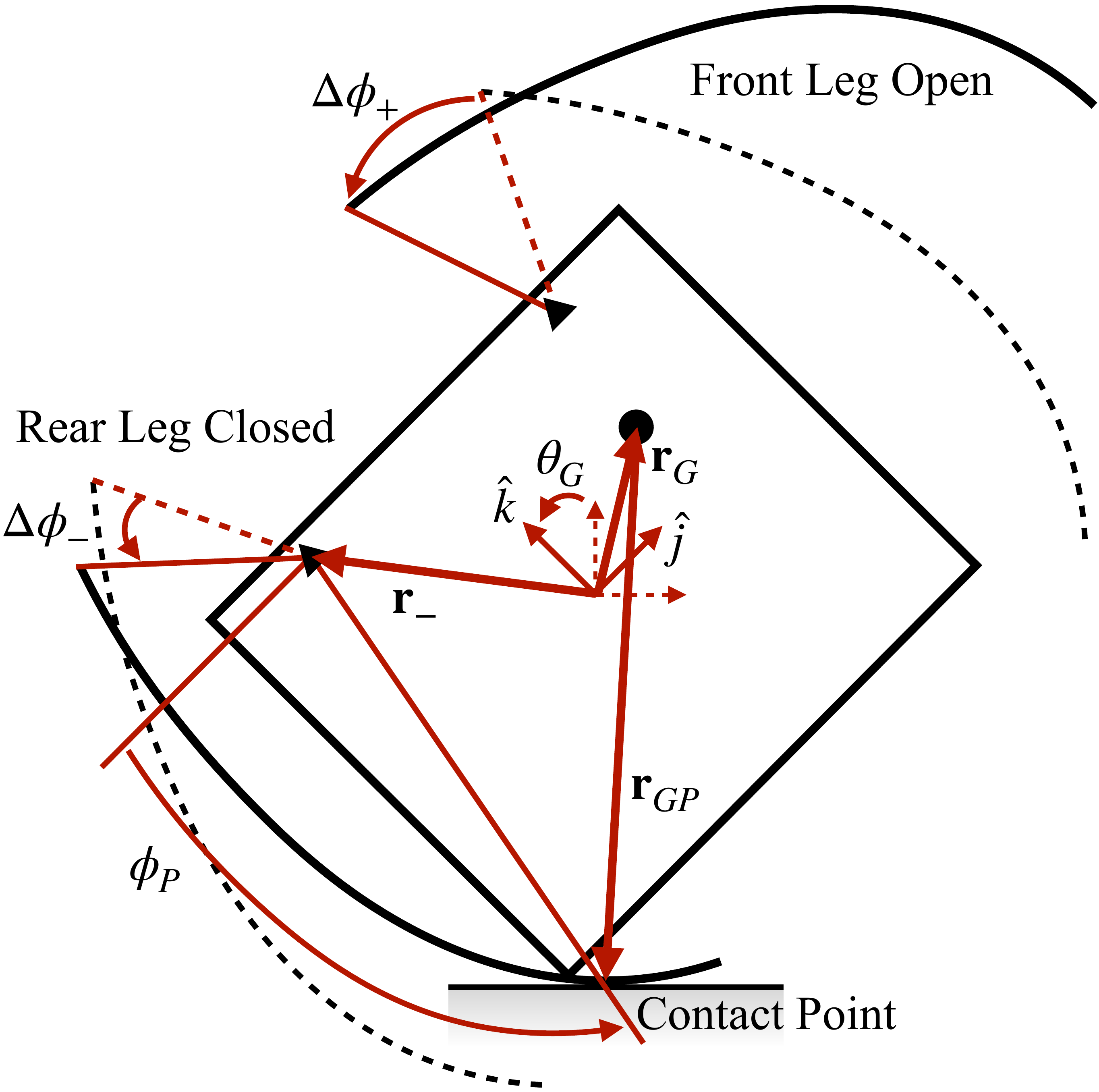}\label{fig:pb}}
\caption{Design parameterization. The origin of the robot's coordinate frame is located at the geometric center of the body. $\hat{j}$,  $\hat{k}$, $\hat{J}$,  $\hat{K}$ are the local forward, local vertical, global forward, global vertical coordinates, respectively. $r(\phi)$ and $\phi_{min} < \phi < \phi_{max}$ are leg radius and angles measured from the rotational joints. Subscripts $+$ and $-$ are used to denote the front and rear legs. The lengths $y_G$, $z_G$, $y_L$, and $z_L$ are vertical or horizontal distances from origin to CoM or rotational joints. \hl{The robot in a rotation of $\theta_G$ is shown in (b), with its rear leg closed ($\Delta \phi_- < 0$) and front leg open ($\Delta \phi_+ > 0$). $\mathbf{r}_G$ and $\mathbf{r}_-$ are position vectors of CoM and rear rotational joint. $\mathbf{r}_{GP}$ is the position vector from CoM to the instantaneous point of ground contact, and $\phi_P$ is the angle measured around the leg to the contact point. }}
\label{fig:dimensions}
\end{figure}

Robot design is begun by defining a  nominal rigid body volume of known dimensions.   Geometric design parameters are then introduced, as summarized in Fig.\,\ref{fig:dimensions}.  We treat the geometric center of the body as the origin for the robot's coordinate frame, with $\hat{i}$, $\hat{j}$, and $\hat{k}$ the local lateral, forward, and vertical coordinates, and $\hat{I}$,$\hat{J}$,$\hat{K}$ the corresponding global ones.  The robot's center of mass (CoM) may be off-center, lying within the body at location $\mathbf{r}_G = y_G \hat{j} + z_G\hat{k}$. Two symmetric legs are attached to rotational joints at positions $\mathbf{r}_{+,-} = \pm y_L\hat{j} + z_L\hat{k}$.  Subscripts $+$ and $-$ denote front and rear legs.   The legs may have a non-uniform radius as measured from the rotational joints, described by function $r(\phi)$, with $\phi$ being angle around the leg as measured from the joint, $\phi_{min} < \phi < \phi_{max}$.   Actuated leg rotation, about each joint, is denoted by $\Delta \phi_{+,-}$.

During rolling on flat ground, for any rotation of the body, $\theta_G$, about the $\hat{i}$ axis, a gravitational moment arm $r_{GP}(\theta_G,\Delta \phi_{+/-})$ can be computed between the CoM and the instantaneous point of contact between either leg and ground. The moment arm can be calculated as:
\begin{equation}
r_{GP}= \left( \mathbf{R} \left( \theta_G \right) \mathbf{r_{+,-}} + \mathbf{R} \left( \theta_G + \Delta \phi_{+,-} + \phi_P \right) \mathbf{r}(\phi_P) \right) \cdot \hat{J}
\end{equation}
\noindent where $\mathbf{R}$ is a standard rotation matrix and $\phi_P$ is the angle measured around the leg to the instantaneous point of ground contact.  The ground contact point is found through a search for the lowest point along the outside of the leg at a given body angle and leg displacement angle.

We hypothesize that rolling locomotion can be achieved with low CoT and small actuator stroke if the cumulative gravitational moment over a complete body rotation can be made large while changes in leg angle are kept small.   Thus, to characterize designs parameterized by $y_G$, $z_G$, $y_L$, $z_L$, and $r(\phi)$, two objective functions are introduced.  The first, $J_1$, evaluates the cumulative moment arm due to gravity during a full rotation of the robot body:
\begin{equation}
    J_1 = \int_{-\pi}^{\pi} \min_{\Delta \phi} r_{GP}(\theta_G,\Delta \phi) d\theta_G
\end{equation}
This minimization assumes the most favorable leg position can be selected for all body angles, meaning a $\Delta \phi$ that makes $r_{GP}$ as negative as possible, for the largest gravitational contribution to clockwise rotation (or smallest barrier, at body angles where a negative $r_{GP}$ is not feasible).  

A second objective function, $J_2$, serves as a proxy for actuation effort. It records the cumulative, absolute change in rotation angle of the legs during a full, idealized body rotation.  This is calculated from the cumulative absolute variation in $\Delta \phi$ that minimizes (2), or:
\begin{equation}
    J_2 = \int_{-\pi}^{\pi} \left| \frac{d}{d\theta_G} \arg \min_{\Delta \phi} r_{GP}(\theta_G,\Delta \phi) \right| d\theta_G
\end{equation}
Objective functions $J_1$ and $J_2$ will not correspond to exact contributions of gravity or actuator effort for rolling.  Rather, they are intended to serve as simple metrics for quickly evaluating many possible designs.  As such, they neglect the dynamics of robot motion and \hl{the fact that actuation effort will depend on the loads on the leg.}  \hl{For $J_1$ to exactly describe gravitational contributions, the robot would need to be able to adjust leg angles perfectly while body angular velocity was constant.  For $J_2$ to exactly describe actuation cost, actuator energy consumption would have to be proportional only to displacement, without dependence on speed or load forces.  These exact conditions are not physically realizable, but the proposed metrics more closely approximate these phenomena when: (1) angular velocity can be kept approximately constant; (2) actuation cost is dominated by whether actuators are active, with minimal dependence on load torque.  The latter can occur in transducers with large torque capacity and large electrical energy dissipation upon activation, which will also tend to mean that leg angles can be changed rapidly}.

\subsection{Design Trends}
Candidate robot geometries were generated by a brute force search over randomized parameters for leg geometry, leg joint location, and CoM location, followed by evaluation of $J_1$ and $J_2$.  Leg geometries $r(\phi)$ were generated within design restrictions so that legs terminate outside the robot body and form curves with \hl{a smooth, spline profile.} Several keypoints were chosen with each keypoint restricted to a specified area relative to the robot body. Then a curve was fit between keypoints to form $r(\phi)$. \hl{Algorithm \ref{alg:leg} shows front leg geometry generation.} During numerical evaluation of $J_1$ and $J_2$, body rotations are calculated by summation over discretized body rotation angles between $-\pi$ and $\pi$. A nominal 60~mm by 60~mm body size was analyzed based on convenient electronic and power components for prototyping.
\hl{
\begin{algorithm}[!t]
\caption{Front Leg Geometry Generation}
\begin{algorithmic}[1]
\renewcommand{\algorithmicrequire}{\textbf{Input:}}
\renewcommand{\algorithmicensure}{\textbf{Output:}}
\REQUIRE Robot body width $w$, height $h$, position of leg joint L
\ENSURE  Front leg geometry $r(\phi)$ 
 \STATE Construct an xy-coordinate system from the bottom, center point of the robot body
 \STATE Randomly choose Point A such that $x_A = 0$, $h < y_A < 2h$ (a starting point above the body)
 \STATE Locate Point B at $x_B = x_A + \epsilon$, $y_B = y_A$, $\epsilon \to 0$ 
 \STATE Rotate $\overline{\rm{AB}}$ around joint L such that $\overline{\rm{AL}}$ is vertical (using Point B an incremental distance from A, with this rotation provides leg tip slope that will form a continuous curve when closed against a symmetric opposing leg)
 \STATE Repeat from Step 2 if $\overline{\rm{AB}}$ intercepts the robot body
 \STATE Randomly choose Points C, D, E, and F (Four random, ordered keypoints outside the body), such that 
 \STATE $x_C > w/2$, $h < y_C < y_B$, $m_{\overline{\rm{BC}}} < m_{\overline{\rm{AB}}}$ 
 \STATE $x_D > x_C$, $w/2 < y_D < h$, $m_{\overline{\rm{CD}}} < m_{\overline{\rm{BC}}}$
 \STATE $w/2 < x_E < x_D$, $y_E < 0$
 \STATE $0 < x_F < w/2$, $y_F < y_E$, $m_{\overline{\rm{EF}}} < m_{\overline{\rm{ED}}}$ 
 \STATE Connect Points A to F using a smoothing spline $r(\phi)$ 
 \STATE Repeat from Step 2 if gradient of $r(\phi)$ is not monotonic
\RETURN $r(\phi)$ 
\end{algorithmic} 
\label{alg:leg}
\end{algorithm}
}
\begin{figure}[!t]
\centering
\includegraphics[width=1.0\linewidth]{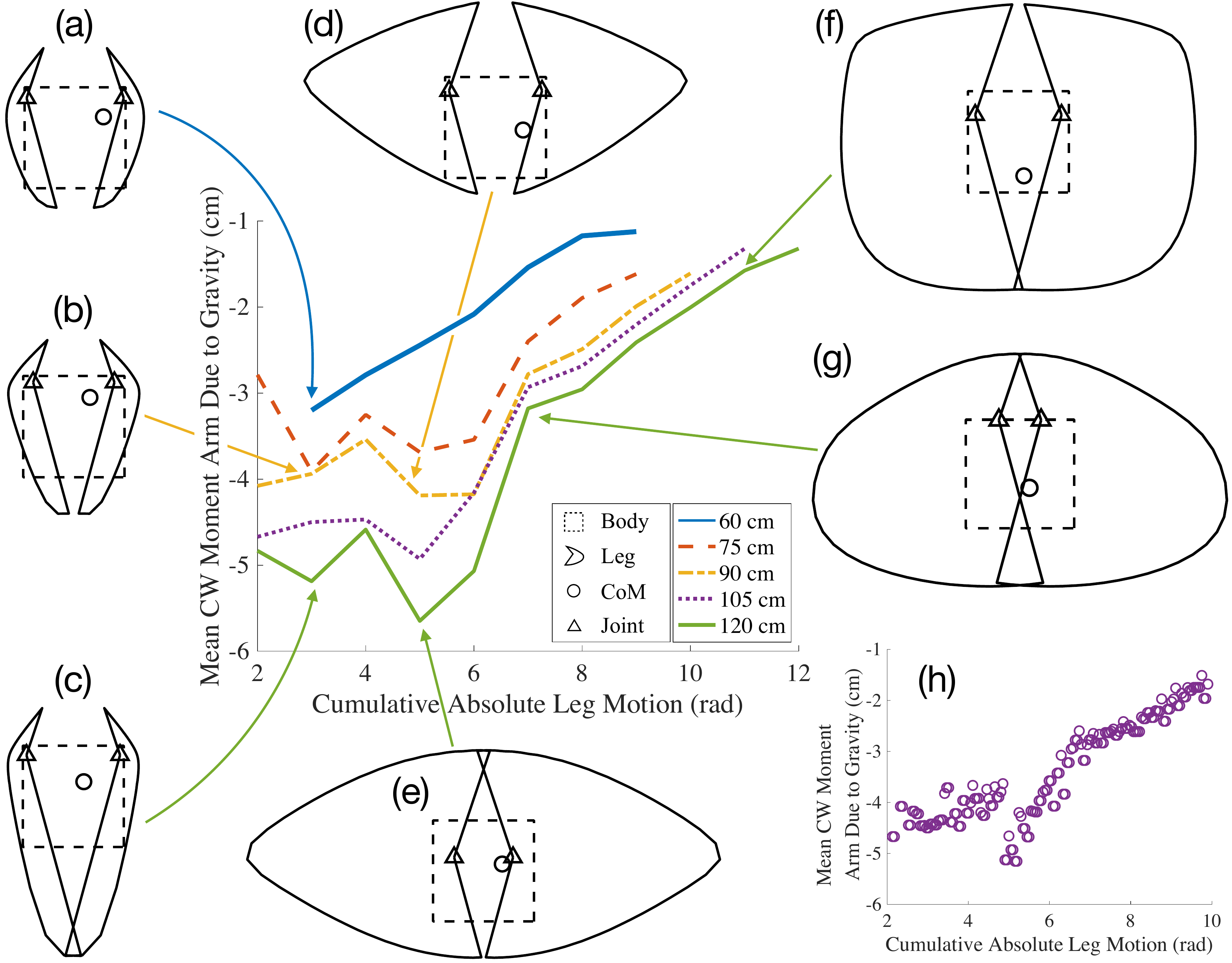}
\caption{Representative design trends by maximum leg radius and cumulative gravitational contribution $J_1$ versus absolute motor rotation $J_2$.  Designs with both small leg motion requirements and large negative gravitational rotation take on an inverted pendulum-like geometry, as in (a), (b), and (c).  Minimal $J_1$ values, as in (d) and (e), were associated with oblong designs where hinged legs primarily manipulate gravitational moment \hl{when the robot is on its ``side"}.  Designs with poor rolling geometry, as in (f) and (g), perform badly by both $J_1$ and $J_2$ metrics.}
\label{fig:Sample_Designs}
\end{figure}

While evaluating candidate leg geometries, several trends were observed.   First, the idealized contribution of gravity to rolling will increase approximately proportionally with leg radius, and thus trends are grouped by maximum leg dimension.   Next, it would be expected that more active leg actuation (larger changes in $\Delta \phi$) will produce more favorable gravitational contribution when other effects are neglected.  However, larger changes in leg angle are also required for ``poor" designs, or leg geometries that are not conducive to smooth rolling.  As a result, after an initial decrease in smallest $J_1$ with respect to $J_2$, $J_1$ begins increasing again.  \hl{Designs in the region where $J_1$ decreases while $J_2$ increases (approx. $J_2 < 5$) can be visualized as a Pareto front that is roughly captured from the random design generation; a detailed plot of design instances along the lower bound of $J_1$ versus $J_2$ is shown in Fig.\,\ref{fig:Sample_Designs}(h), for 105~mm leg length}.

Examining leg shapes produced by the proposed evaluation metrics, three main types of geometry were observed.   When using the smallest actuator efforts (minimizing $J_2$), the metrics select for tall, narrow leg geometries, such as sample designs (a), (b), and (c) in Fig.\,\ref{fig:Sample_Designs}. Leg connection points are placed at the upper corners of the body and center-of-mass placed modestly off center.   The resulting designs resemble an imbalanced inverted pendulum when upright with legs closed, then rely on legs opening as the body rotates to achieve smooth rolling.   A less intuitive result arises for designs with the largest gravitational contribution, illustrated in Fig.\,\ref{fig:Sample_Designs}(d)-(e).   These designs select for elongated leg geometries that primarily allow movement of the legs to shift center of the mass relative to contact point when the robot is on its side (i.e. $\theta_G \approx \pm \pi$/2).  Practically, with such leg geometries it would be difficult to initiate rolling from the upright position and to coordinate leg rotation with sufficient accuracy to avoid a large barrier to rotation near $\pm \pi/2$, illustrating some limitations of the quasi-static analysis.  Finally, leg designs with largest $J_2$ tend to reflect poor rolling geometry by either metric, as in sample designs \hl{(f) and (g)}, and also provide worse (less negative) values for $J_1$. 

The conclusion of this analysis was that gradually increasing radius from the upper to lower part of the legs, resulting in an inverted pendulum-like geometry with legs closed, might be promising for efficient rolling \hl{(best $J_1$ for the lowest cost of $J_2$)}.    This geometry was selected for further prototyping and testing.   The chosen design corresponds nearly to design (b) from Fig.\,\ref{fig:Sample_Designs}, with minor adjustments to simplify manufacturing.   Total range of motion required was 55 degrees.

\section{Controller Design}
\subsection{Control Concept}
The leg design procedure also provides guidance for feedback controller development. Fig.\,\ref{fig:contour} shows a contour plot for the gravitational moment arm acting on the robot's CoM, as a function of both rotation angle of the body (free to rotate over a full $2\pi$ rad in the coordinate frame) and leg position (able to rotate approximately -10 to +45 degrees from its nominal position relative to the body).  The independent control input is the leg position.  For forward rolling in the defined coordinate axes, negative gravitational moments \hl{(darker regions in Fig.\,\ref{fig:contour})} are desirable, corresponding to clockwise (CW) torque.  Thus, identifying the leg angle for each body angle providing the most negative gravitational moment is used as a guide for regulating leg positions.  

Within Fig.\,\ref{fig:contour}, we show an idealized leg trajectory, where the robot's gravitational moment is made as negative as possible at each body orientation. This trajectory also largely matches intuition about robot motion. When the robot body is upright, the gravitational moment is negative if the legs close, resulting in a natural ``fall" to the right.   As the robot rolls onto the front leg, near a $-\pi/2$ rotation, it becomes advantageous to gradually open the legs, keeping the CoM to the right of the leg contact point for as long as possible. With the legs fully open, a smooth, continuous connection between the legs is presented as the body rolls past $-\pi$.  Finally, the rear leg closes as the body rotates from \hl{$\pi$} to approximately \hl{$\pi/2$}, at which point the robot relies on its momentum to return to an upright position and repeat the cycle.  \hl{Legs should be able to rotate at least as fast, relative to the body, as the maximum slope in Fig.\,\ref{fig:contour}: in this case, 55$^\circ$ over $\pi/3$ rad or 60$^\circ$ of body rotation.}

As noted earlier, this formulation neglects actuator work done, which can be large when the robot is opening its front leg.  Nonetheless, abstracting target leg angle to a function of body orientation greatly simplifies sensor feedback required to maintain a whole-body rolling gait, while allowing passive dynamics to contribute significantly to rolling locomotion.   

\subsection{Event-drive Implementation}

\begin{figure}[!t]
\centering
\includegraphics[width=1\linewidth]{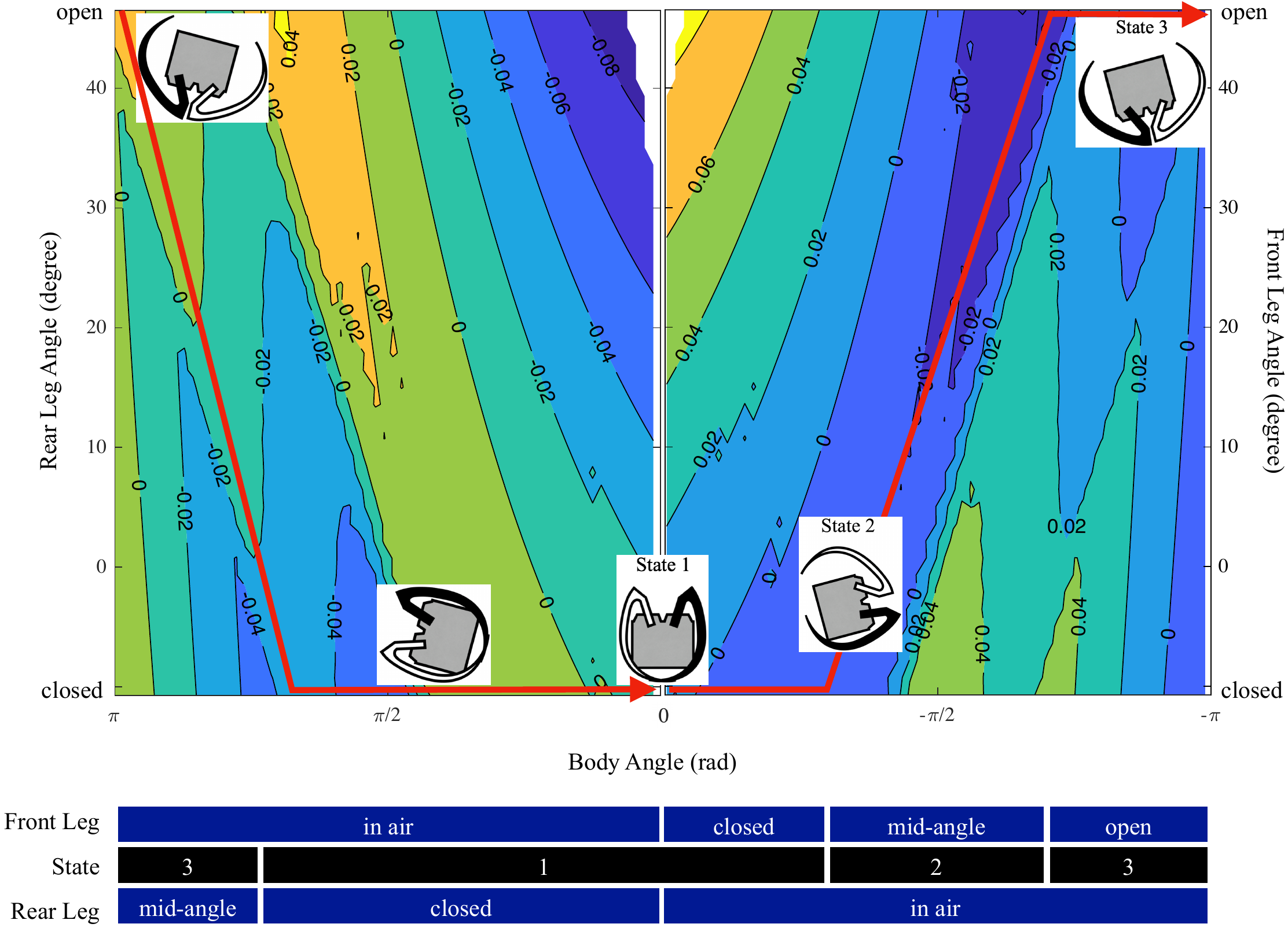}
\caption{Leg control abstraction to body rotation state. Contour plots show gravitational moment arm versus body angle and leg angle for front (\hl{black}) and rear (\hl{white}) legs, with body-leg state trajectory marked that minimizes gravitational moment at each orientation, to contribute negative torque to clockwise (CW) forward rolling. Note: x-axis is plotted from positive to negative to allow negative rotation to be illustrated moving to the right.  \hl{Contour bars indicate gravitational moment arm, with darker regions more favorable to CW rolling, and lighter regions being barriers to CW rolling.}}
\label{fig:contour}
\end{figure}

In implementation, to simplify control and improve robustness against sensor noise, the idealized leg trajectories are divided into \hl{discrete regions with widths comparable to the angular resolution of the sensor measuring body rotation.  In our prototype robot using a 3-axis accelerometer, we estimated our angular resolution at $\pm 15 ^\circ$, and thus applied discretized leg positions covering not less than $\pi/6$ rad.  This resulted in either open-, closed-, or mid-angle leg positions that would be desirable to approximate the nominal continuous trajectory.}   The resulting event-driven controller for rolling has five states: three of these correspond to major regions of the contour plot in \hl{Fig.\,\ref{fig:contour}}, and are summarized below the contour plots;  two others are recovery states, added to assist if the robot fails to complete a roll. \hl{Conceptually, closer tracking of the idealized leg trajectory could be obtained with high sensor resolution (by reducing noise and/or increasing sampling rate) or use of an estimator for body angle and velocity.}  
\begin{figure}[t]
\centering
\includegraphics[width=1\linewidth]{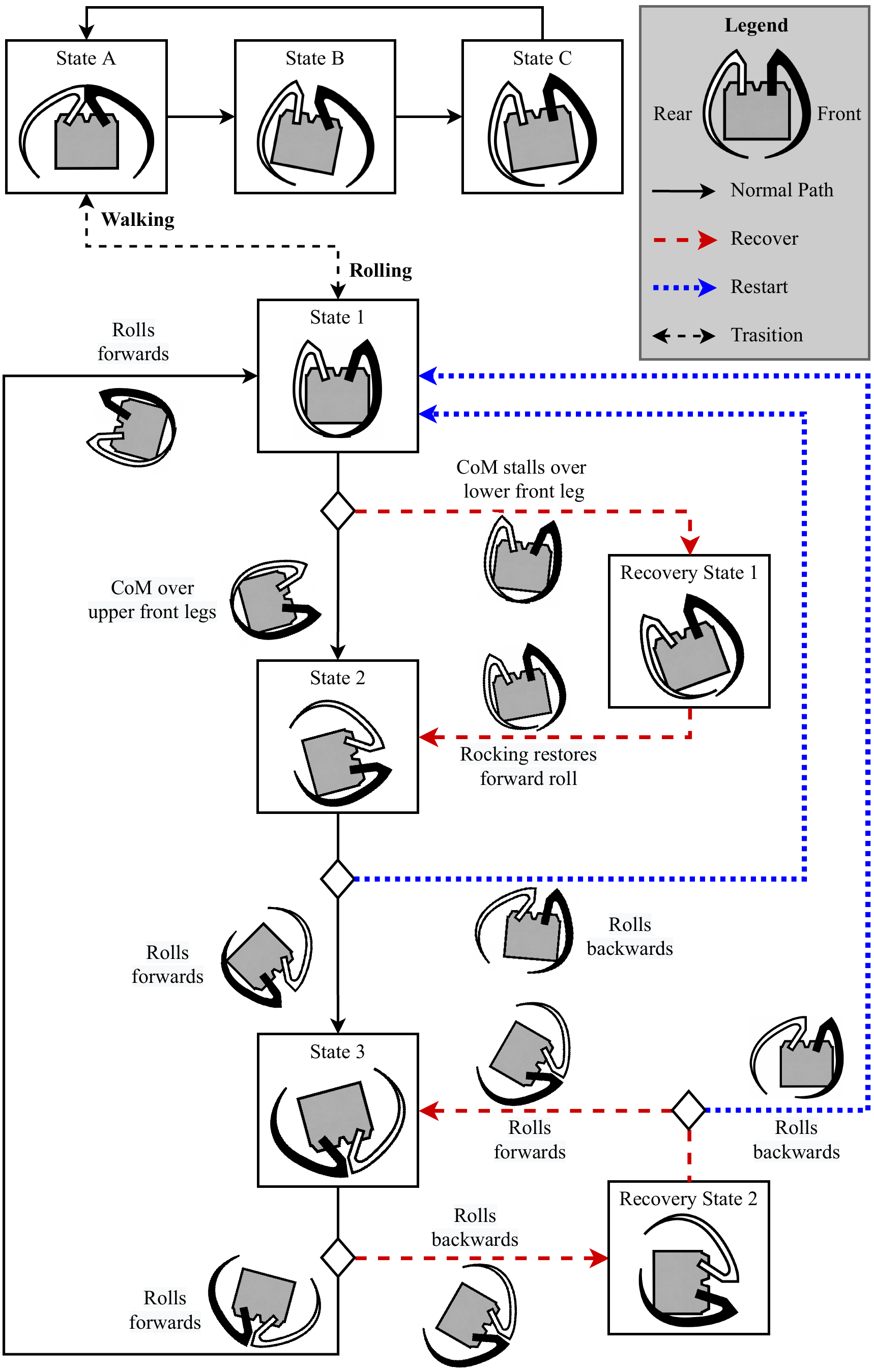}
\caption{Control diagram. States A, B, and C are a walking controller that allow the robot to walk continuously.  States 1-\hl{3} are used by the rolling controller.  Transition between walking and rolling is performed by opening or retracting both legs when the robot is upright in State A or State 1. While rolling, states 1, 2, and 3 allow the robot to roll forward continuously. Recovery states \hl{1} and \hl{2} form additional steps in restarting sequences after interruptions.}
\label{fig:control-diagram}
\end{figure}

The event transitions and corresponding body-angle and leg-angle states for rolling are shown in Fig.\,\ref{fig:control-diagram}, accompanied by a \hl{connected} sequence for walking locomotion.    A standard rolling sequence begins from an upright orientation with legs closed (State 1), from which the robot will rotate under gravity onto its front leg.  The front leg is then partially opened (State 2) to achieve a approximation of the idealized leg angle vs. body angle trajectory. Target leg angle and body orientation thresholds defining State 2 are selected to maximize the range of body angles over which the gravitational moment remains negative.  While CoM is over upper front legs, the legs open completely (State 3), to be ready to roll past $-\pi$ body angle.   Finally, once the robot orientation passes $-\pi$, both legs are closed again (State 1).   While the sequence corresponds to the intuitive understanding of the robot's rolling gait, the gravitational moment arm analysis aids in selecting desirable body angles at which each state transition occurs.   

Two recovery states were designed by trial-and-error for situations where the robot does not complete the normal rolling steps. These are used if the robot rolls backwards at any time, generally due to irregularities of the ground (see Section V.E) and imperfect matching of the legs where they meet in the physical prototype.    If rollback is detected or the CoM stalls over the lower front legs in State~1, Recovery State 1 is triggered, with the intention of using the rear leg to re-initiate forward rolling. If the robot rolls backward in State 3, both legs are closed (State 1) to restart the rolling sequence.  Should the robot fail to generate sufficient momentum to cross -$\pi$~rad body angle in State 3, Recovery State 2 retracts the front leg while the rear leg remains open; if body orientation remains near -$\pi$~rad, with CoM still over the upper part of the leg, the robot returns to State 3, during which the rapid opening of the front leg may renew rolling motion.  Otherwise, the robot returns to State 1 to restart the entire sequence.

\subsection{Walking Control and Transitions}
A simple control sequence was also designed for a two-legged walking gait. This control sequence is inspired by inchworm locomotion, but must maintain sufficient normal force on the rear leg to avoid slipping given the robot's forward-shifted CoM.  From standing (State A), the robot first retracts its back legs to lean its body backwards (State B). Then, the robot partially retracts its front legs (State C), so that the COM is over its back legs to maximize normal force at their contact point. The robot then returns to State A, rapidly opening the back legs and kicking the body forward, while opening its front legs to prepare for touchdown. 

Walking and rolling gaits are connected through State A and State 1, during both of which the robot is standing upright. By either opening (State A) or closing (State 1) both legs in this orientation, the robot can transit between walking and rolling leg actuation sequences. 
\begin{figure}[!t]
\centering
\subfloat[]{\includegraphics[width=0.48\linewidth]{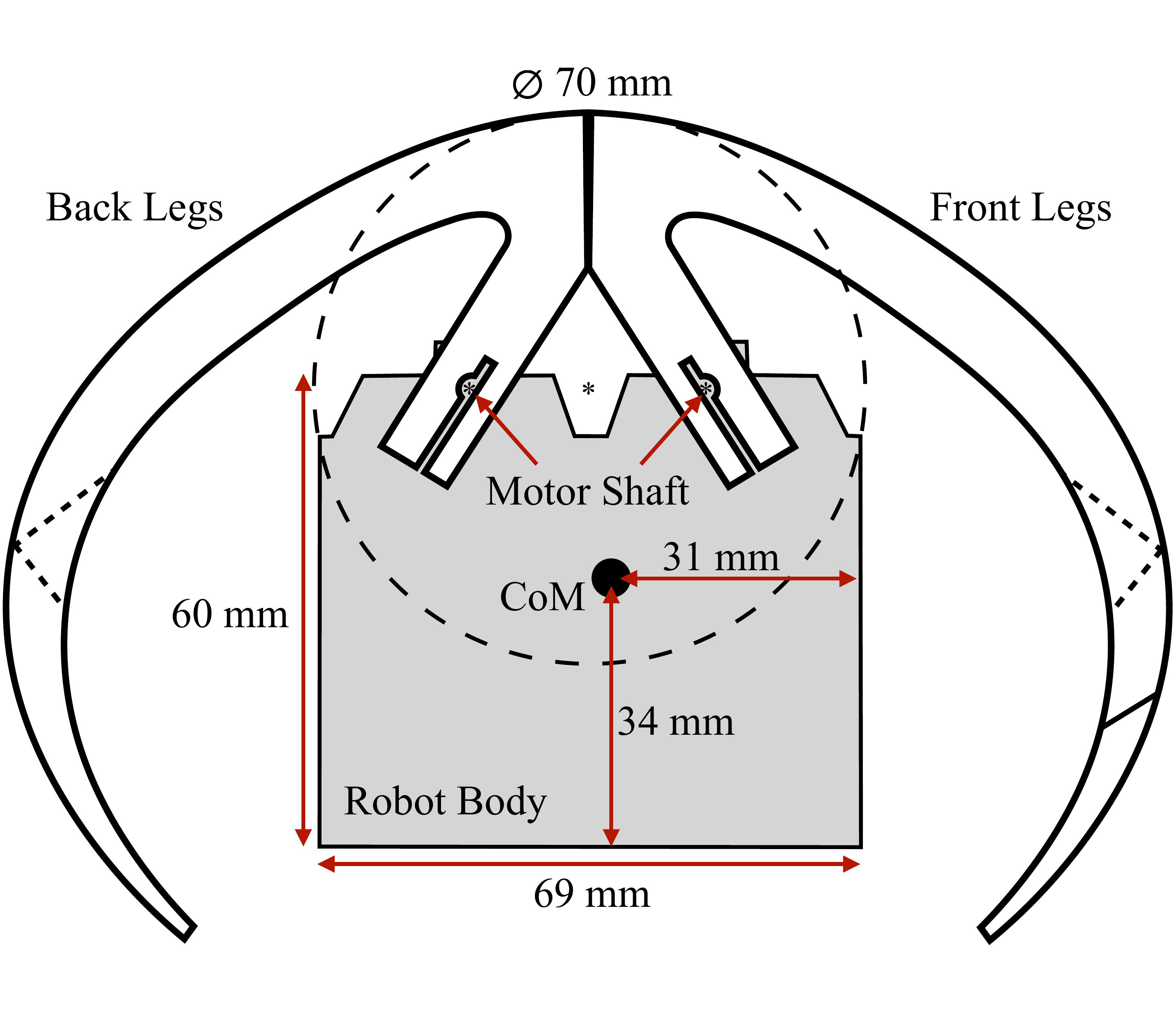}\label{fig:3a}}
\hfil
\subfloat[]{\includegraphics[width=0.52\linewidth]{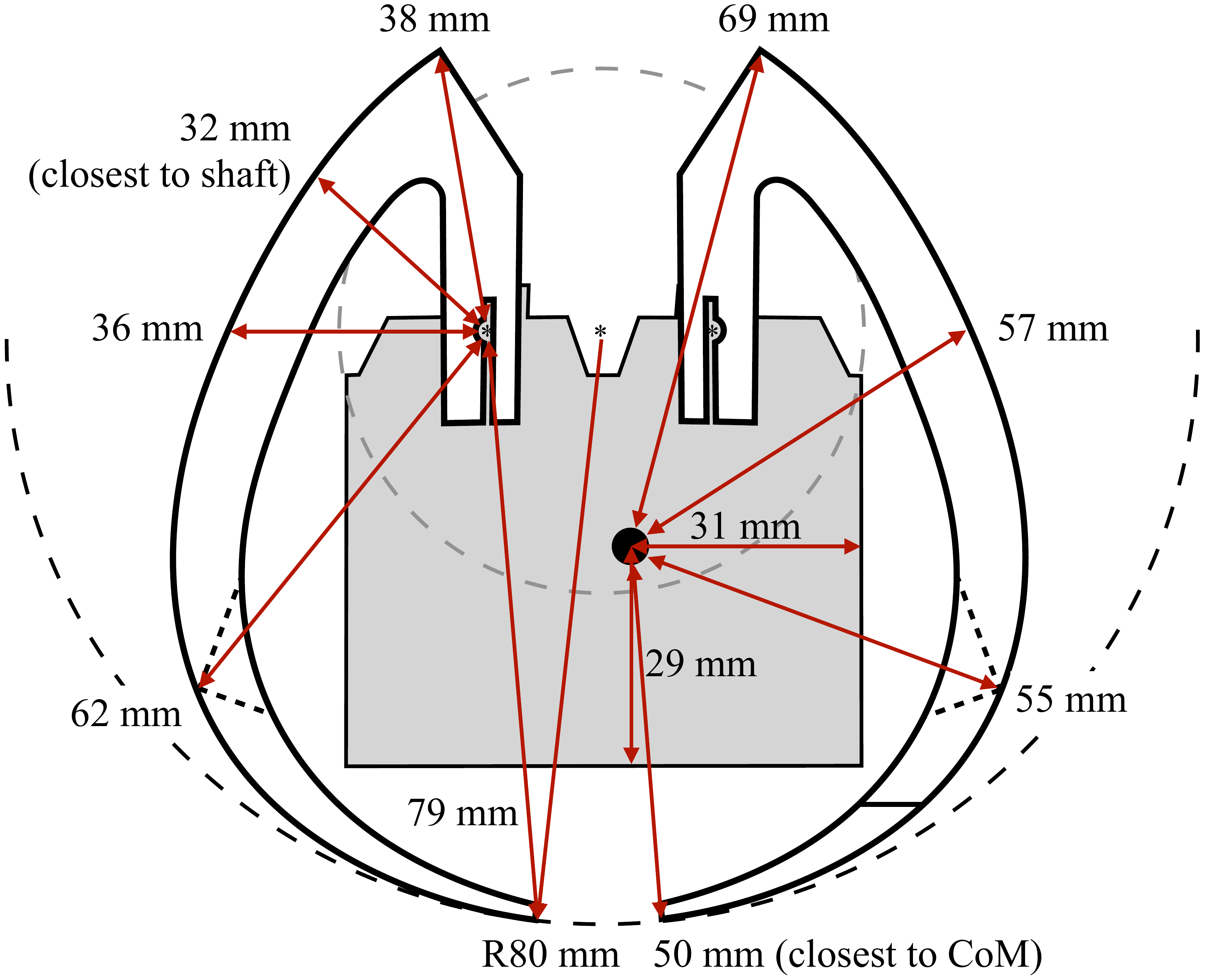}\label{fig:3b}}
\caption{Leg profile. The positions of the center of mass and motor shafts are indicated in (a). When the robot opens its legs to the maximum angle, the legs form a circle with a diameter of 70 mm and a center at the midpoint between two motor shafts, placing the robot's center of mass at its lowest point when the robot upside down. The robot's upright stance is shown (b). The legs are smooth curves with varying radius, with minimum distance of 32 mm to the motor shaft and \hl{50} mm to the center of mass.}
\label{fig:leg-profile}
\end{figure}

\section{Prototyping}
The multi-modal robot design and control structure were tested as a meso-scale prototype, to evaluate walking and rolling performance with the proposed curved leg geometry.

\subsection{Fabrication and Assembly}
As a testbed for untethered multi-modal gait, robot size was chosen to accommodate an Arduino Uno microcontroller and two 3.7 V rechargeable batteries (icr18650). The prototype is actuated by four gearmotors (DFRobot FIT0487 DC Geared Motor), mounted in pairs on top of the chassis.  \hl{Two motors acting synchronously on each leg were calculated as necessary to provide sufficient torque to hold static leg positions when upright}. In addition, while capable of continuous rotation, in this setting the gearmotors represent high-force, finite-stroke transducers, and were limited to a range-of-motion of 55 degrees.   \hl{Robot CoM was matched to the design target as closely as possible through selection of the battery pack location in the chassis, as it has sufficient mass relative to the rest of the robot to alter CoM position substantially.}   

The robot’s electrical system consists of a microcontroller, a motor driver, and interface electronics. An Arduino UNO Wifi Rev 2 supports encoders for each of the actuation motors and performs local PID motor control and event-driven gait control.  Each of the motors is connected to a motor driver, which is sourced from the battery directly. The motor driver is supplied with control commands via pulse-width modulation (PWM) from the microcontroller.

Leg profiles as manufactured are shown in detail in Fig.\,\ref{fig:leg-profile}, along with the off-center CoM and location of leg axes of rotation and effective rolling radii in open and closed leg positions.  \hl{The empty chassis (62~grams), front legs (62~grams), and rear legs (56~grams)} are manufactured by 3D printing using ABS P430.  \hl{To reduce leg weight, only the outer edges of the legs feature the continuous design profile, while the center portion uses cross-bracing with outside contour matched to the legs.   This means that on uneven terrain, forces may not remain evenly distributed across the lateral axis of the robot, but these 3D effects have not been considered in the current work.}  With microcontroller, motors, and battery installed, total robot mass is $373$~grams. \hl{In its upright stance (Fig.\,\ref{fig:leg-profile}(b)), the robot has dimensions (length $\times$ height $\times$ depth) of 110~$\times$~113~$\times$~109 mm$^3$.}  \hl{While leg mass was neglected in the quasi-static analysis, we calculate open- and closed-leg positions to alter CoM location by at most 20~mm.  The largest change occurs during State 2 and is primarily vertical, thus having little effect on gravitational contribution to rolling}. 

\subsection{Sensing and Feedback}
Feedback available to the microcontroller consists of measurements from encoders at both legs and from inertial sensor in the body.  Leg rotations are measured by front and back encoders. Body accelerations and angular velocity are measured by an inertial measurement unit (IMU) on the Arduino board. Gravitational acceleration present in the acceleration measurement is used to approximate body angle from longitude and vertical acceleration.  The IMU has limited sampling rate and poor signal-to-noise ratio at the acceleration and angular velocity magnitudes achievable by the robot, but this is considered representative of limitations that a robot would face under size and power constraints at even smaller scales.

Gait control is implemented in a multi-loop structure.  The \hl{outer} loop uses the accelerometer feedback to measure body orientation, from which target leg angles are selected by the event-driven controller of Section III.  The inner-loop is a PID controller that regulates leg position to target angles using encoder measurements.  Inner-loop regulation also prevents drift in leg position during the periodic walking gait.

\section{Results}
Locomotion was tested under periodic walking control, event-driven control of rolling motion, and during transitions between modes.   Power requirements and CoT are estimated, with discussion of implications and limitations for miniaturization of multi-modal locomotion.

\begin{figure*}[!t]
\centering
\subfloat[]{\includegraphics[width=0.5\linewidth]{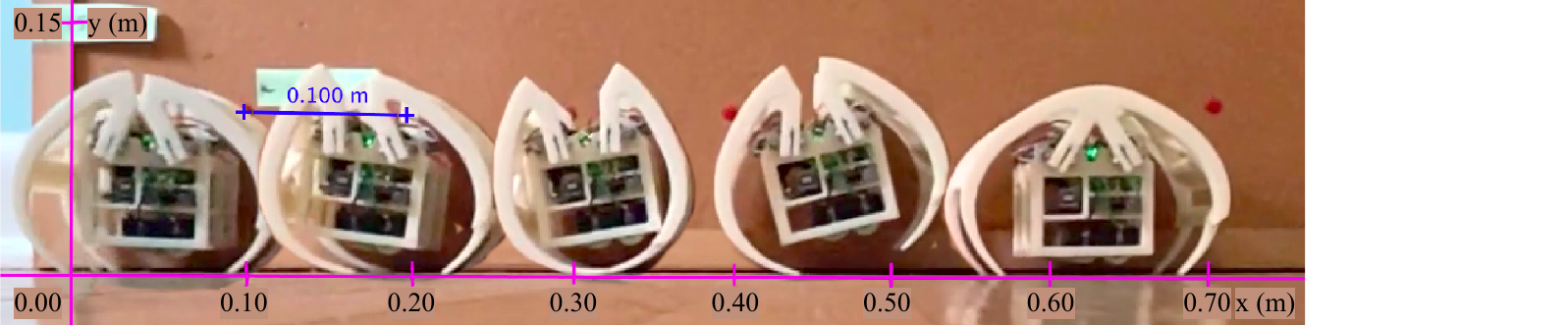}\label{fig:6a}}
\hfil
\subfloat[]{\includegraphics[width=0.5\linewidth]{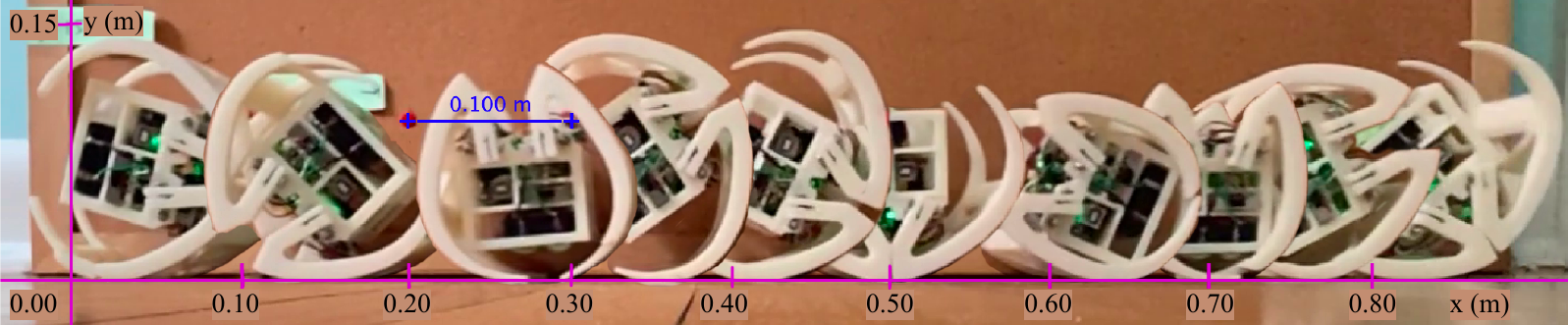}\label{fig:6b}}
\hfil
\subfloat[]{\includegraphics[width=0.49\linewidth]{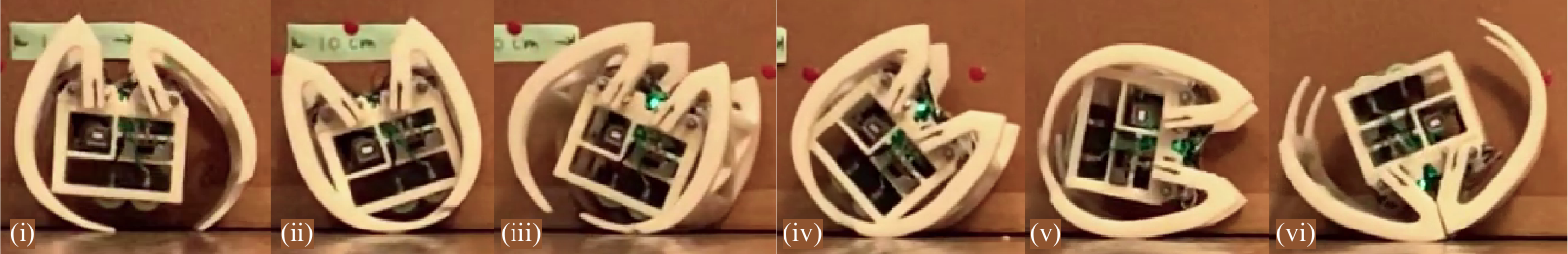}\label{fig:6c}}
\hfil
\subfloat[]{\includegraphics[width=0.5\linewidth]{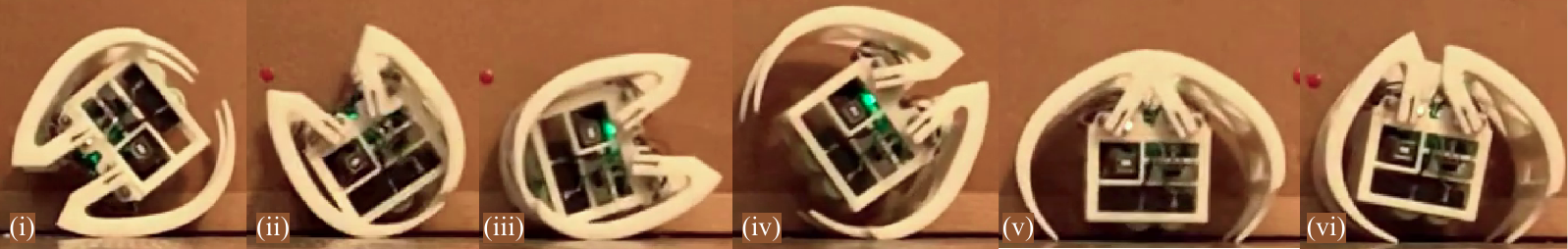}\label{fig:6d}}
\caption{Snapshots of motion during (a) walking, (b) rolling, and (c) transition. (a) A sequence of snapshots of the walking process of the robot, captured in different walking cycles. (b) A sequence of snapshots of the rolling process of the robot, captured in two rolling cycles and 0.5 seconds for starting-up before rolling. (c) Snapshots of states of a walking-to-rolling transition. (d) Snapshots of states of a rolling-to-walking transition.}
\label{fig:snapshots}
\end{figure*}

\begin{figure}[!t]
\includegraphics[width=1\linewidth]{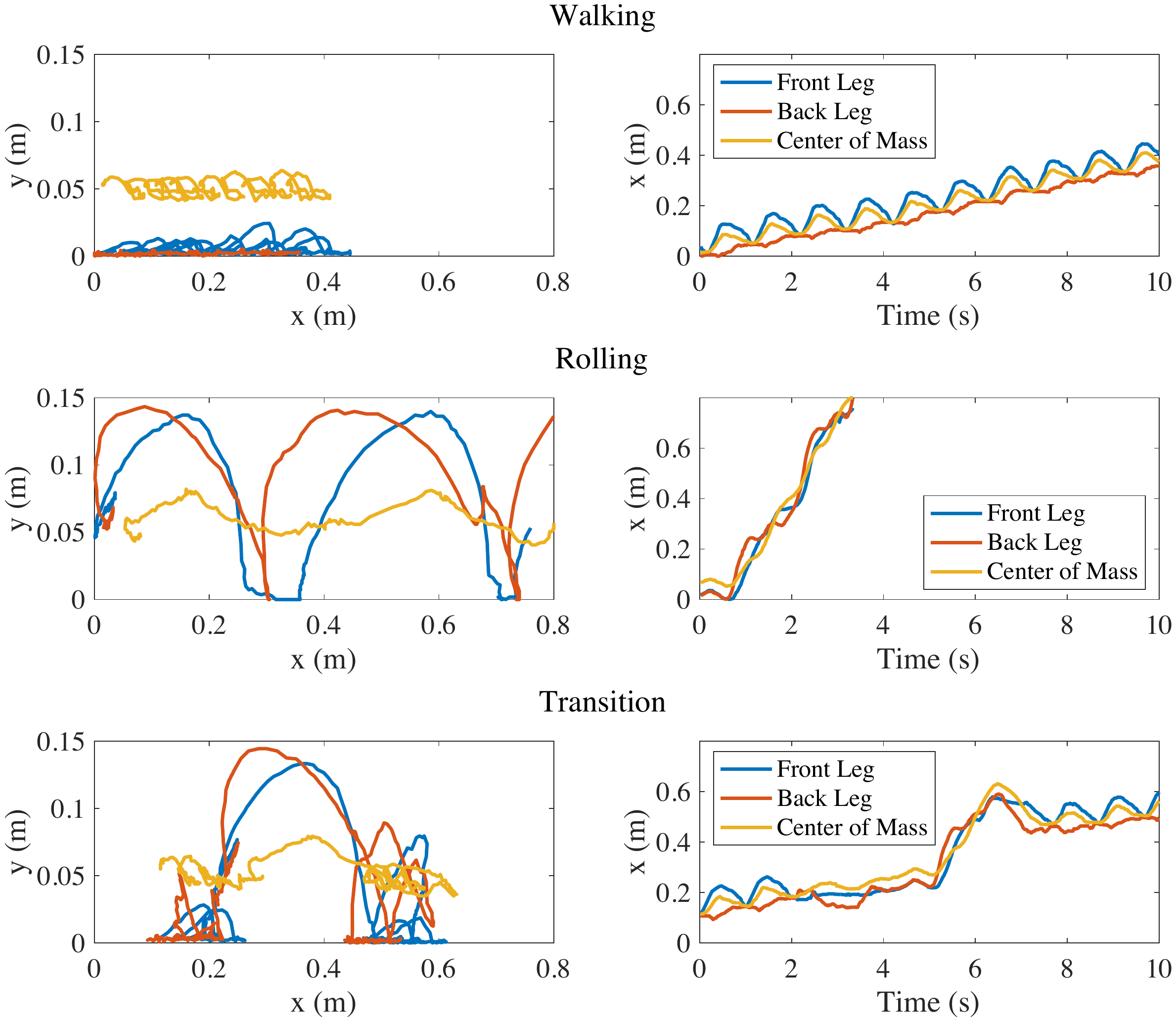}
\caption{Results of motions in walking, rolling, and transition. The x and y locations of the robot’s center of mass, the tip of its front leg, and the tip of its back leg are shown in the \hl{right column}. The x locations vs time in 10 seconds are shown in figures in the \hl{left column}.}
\label{fig:motion-results}
\end{figure}

\subsection{Walking}
Experimentally, the robot achieves a forward speed at approximately 0.038~m/s during walking, and one walking cycle takes about one second.  Fig.\,\ref{fig:snapshots}(a) shows a sequence of snapshots of the walking process of the robot, captured at different moments in one experiment.  The robot requires 10 walking cycles over almost 10 seconds to move forward 0.38 meters. The locations of its center of mass and the tips of the front and back legs were tracked \cite{tracker} and the results are shown in the first row of Fig.\,\ref{fig:motion-results}. Walkign speed is limited by the need for the robot to re-position its CoM relative to the rear-leg during each walking cycle to avoid excessiv slip while pushing forward.  The rapid opening of legs before touchdown also raises a risk of large vibrations or bouncing at impact, but this was not observed experimentally, perhaps due to compliance and damping losses at the  leg and gearmotor connections.

\subsection{Rolling}
Experimentally, the robot rolls forward continuously on level ground using the event-driven controller.  Fig.\,\ref{fig:snapshots}(b) shows a sequence of snapshots of the rolling process, capturing two rolling cycles and 0.6 seconds of start-up before rolling.  In this sequence, the robot completes approximately two rolling cycles over 2.5 seconds, moving forward 0.75 meters. The trajectories of the CoM and the tips of the legs are shown in the middle row of Fig.\,\ref{fig:motion-results}.  Sustained rolling locomotion on level ground has an average speed of 0.24~m/s, and is observed to be both smoother and more efficient than legged locomotion, with the CoM continuously moving forward.

\subsection{Walking-to-Rolling Transition}
The robot can also smoothly change between walking and rolling mode. Fig.\,\ref{fig:snapshots}(c) and Fig.\,\ref{fig:snapshots}(d) show sequences of snapshots of the transition from walking to rolling and the transition from rolling to walking. Snapshots are captured in one transition process. The transition from walking to rolling is analogous to the start-up stage of rolling, and the robot maintains forward velocity throughout. On the other hand, the robot stops moving forward during the transition from rolling to walking, as it needs to halt momentum to stand upright. That step back action is can be observed in Fig.\,\ref{fig:snapshots}(d)(iv).  Robot motion during transitions is shown in the \hl{the last row of} Fig.\,\ref{fig:motion-results}. 

\begin{table*}[!t]
\caption{Speed and cost-of-transport (when available) of small multi-modal robots in legged and rolling locomotion as reported in literature, with select comparative single-mode robots.}
\centering
\renewcommand{\arraystretch}{1.3}
\begin{threeparttable}
\begin{tabular}{llcc|ccccc |ccccc}
\toprule
& & & \multicolumn{1}{c}{} & \multicolumn{5}{c}{\textbf{Rolling Gait}} & \multicolumn{5}{c}{\textbf{Legged Gait}} \\
\midrule
 &  & \textbf{Mass} & \textbf{Length,} & & \textbf{Speed} & \textbf{Speed} & \textbf{Power} & & & \textbf{Speed} & \textbf{Speed} & \textbf{Power} \\
 \\[-1.8em]
\multirow{-2}{*}{\textbf{Author}} & \multirow{-2}{*}{\textbf{Transduction\tnote{1}}} & \textbf{(kg)} & \textbf{Body (m)} & \multirow{-2}{*}{\textbf{DoF}} & \textbf{(m/s)} & \textbf{(BL/s)} & \textbf{(w)} & \multirow{-2}{*}{\textbf{CoT}} & \multirow{-2}{*}{\textbf{DoF}} & \textbf{(m/s)} & \textbf{(BL/s)} & \textbf{(w)} & \multirow{-2}{*}{\textbf{CoT}} \\
\midrule
Asiri \cite{Asiri2019} & DC (\textgreater{}360$^{\circ}$) & 6.36 & 0.32 & 2 & 0.64 & 2.00 & 37 & 0.93 & --& \multicolumn{3}{c}{Rolling only}&-- \\
Jia \cite{Jia2017} & DC ($\sim$90$^{\circ}$) & 5.00 & 0.73 & 12 & 0.50 & 0.69 & -- & -- & 12 & 0.12 & 0.15 & -- & --\\
Luneckas \cite{Luneckas2021} & DC (\textgreater{}360$^{\circ}$) & 1.50 & 0.25 & --& \multicolumn{3}{c}{Legged only}&-- & 18 & 0.33 & 1.32 & -- & 5.75 \\
Aoki \cite{Aoki2020} & DC ($\sim$45$^{\circ}$) & 1.30 & 0.19 & 12 & 0.22 & 1.18 & -- & -- & 12 & 0.03 & 0.16 & -- & --\\
Phipps \cite{Phipps2008} & DC (T) ($\sim$90$^{\circ}$) & 1.13 & 0.22 & 3 & 0.06 & 0.26 & 76 & 3.43 & 3 & 0.03 & 0.15 & 670 & 30.3 \\
Miura \cite{Miura2019}\cite{Miura2019-2}& DC ($\sim$30$^{\circ}$) & 0.77 & 0.25 & 18 & 0.25 & 0.33 & -- & -- & 18 & 0.04 & 0.05 & -- & --\\
Nemoto \cite{Nemoto2015} & DC ($\sim$45$^{\circ}$) & 0.39 & 0.21 & 2 & 0.29 & 1.39 & -- & -- & 12 & --  & --  &  -- & --\\
\textbf{This work} & \textbf{DC (55$^{\circ}$)} & \textbf{0.37} & \textbf{0.11} & \textbf{2} & \textbf{0.24} & \textbf{2.24} & \textbf{0.46} & \textbf{0.52} & \textbf{2} & \textbf{0.04} & \textbf{0.37} & \textbf{0.67} & \textbf{4.62} \\
Ho \cite{Ho2013} & DC (\textgreater{}360$^{\circ}$) & 0.1 & 0.18 & 2 & 0.60 & 3.33 & 1.70 & 28.4 & --& \multicolumn{3}{c}{Legless hopping}&-- \\
Li \cite{Li2010} & DC (\textgreater{}360$^{\circ}$) & 0.02 & 0.12 & --& \multicolumn{3}{c}{Legged only}&-- & 6 & 0.80 & 6.67 & -- & 0.82 \\
Duduta \cite{Duduta2020} & DE (T) & 0.01 & 0.05 & 1 & 0.06 & 1.28 & 35 & -- & 1 & 0.02 & 0.40 & -- & 113 \\
Lin \cite{Lin2011} & SM (T) & 0.01 & 0.10 & 2 & 0.63 & 6.25 & 20.7 & 796 & 2 & 0.08 & 0.78 & 10.3 & 2200 \\
Goldberg \cite{Goldberg2018} & PE & \textless{}0.01 & 0.05 & --& \multicolumn{3}{c}{Legged only}&-- & 8 & 0.17 & 3.80 & -- & 29.6 \\
Patel \cite{Patel2018} & PE (T) & \textless{}0.01 & 0.02 & --& \multicolumn{3}{c}{Legged only}& -- & 6 & 0.04 & 1.80 & -- & 3.6 \\
\bottomrule
\end{tabular}
\begin{tablenotes}
\item[1] DC: DC motor, DE: Dielectric elastomer, SM: Shape memory alloy, PE: Piezoelectric, T: Tethered
\end{tablenotes}
\end{threeparttable}
\label{table:benchmark}
\end{table*}



\subsection{Energy and Sensor Usage}
Energy consumed by each motor, $E_i$, is estimated by summing electric power over controller sampling periods for the duration of a given experiment,
\begin{equation}
    E_i = \sum( d V i_{max} T)
\end{equation}
where subscript $i$ indicates the motor (of 4), $d$ is the period's PWM duty cycle, $V$ is battery voltage, $i_{max}$ is motor stall current, and $T$ is controller sampling period.  \hl{Duty cycle is included to quantify the percentage of time that voltage is applied during each time step, while a fixed stall current is used throughout.  Use of stall current produces} a conservative estimate of power consumption: true current draw will be lower when there is a non-zero motor velocity, but this was not directly measurable with existing hardware.

Cost of transport  is computed from estimated battery energy consumed per transit distance times robot weight, 
\begin{equation}
    CoT = \frac{\sum_{i=1}^4 E_i}{mgD}
\end{equation}
where $m$ is robot mass, $g$ is gravitational constant, and $D$ is distance traveled. 

Using recorded PWM commands and the assumption of maximum (stall) current during PWM `on' periods, energy consumption is approximately 16.7 joules per meter during walking (average power 0.67~W)  and 1.9 joules per meter during rolling (average power 0.46~W).   This equates to a CoT of 4.62 during walking and 0.52 during rolling. Speed (in body lengths per second, BL/s) and energy requirements are summarized  in Table~\ref{table:benchmark}, relative to robots for which results have been previously reported at similar size scales.  The robot's CoT in rolling is particularly low for multi-modal locomotion.  While an ideal wheel on level ground has negligible CoT, implementing whole-body robot rolling in practice incurs substantial actuation effort in more complex robot designs. Speed and CoT during walking is inferior to robots designed to optimize running efficiency, but better than most similarly-scaled multi-modal robots. 

The event-drive controller functions effectively in the presence of substantial inertial sensor noise.  Relying on the limited number of discrete ranges of body angle to choose target leg locations reduces sensitivity to large transient impulses in IMU output and limited IMU sampling rate. 

\subsection{Discussion}
The proposed robot design is especially effective in achieving low CoT during whole body rolling.   While CoT of an ideal wheel on a smooth surface after rolling begins is zero, in practice robots designed to roll continuously require substantial actuator inputs, and power consumption can be large when manage multiple actuators in prior designs.   While simplistic, the evaluation of non-uniform-radius, curved leg designs based on their ability to maximize gravitational contribution to rolling results in almost an order-of-magnitude improvement in CoT among multi-modal robots where sufficient data has been published for comparison.   Meanwhile, even with just two DoF, rapid transitions between rolling and walking gaits can be achieved, for improved versatility compared to rolling alone.

\hl{Evaluation of terrain compatibility is less rigorous.  Rolling locomotion was achieved with little variation on surfaces including wood, tile, glass, and low-ply carpet, and over obstacles or irregularities up to 8~mm.  Walking locomotion modestly increases passable terrain: the robot could traverse grades up to 5.1$^\circ$ and clear obstacles up to 12~mm in walking mode, though this remains an area for potential improvement.   Motion is also subject to constraints on surface friction, as at least one leg must be able to slide along the ground during walking Stage B or to close both legs to initiate rolling.  Failure occurred at a coefficient of friction of approximately 0.7 (on a polyvinyl foam mat), upon which legs became stuck open.} 

There are other key limitations.  The rolling gait is unidirectional, \hl{and steering has not been examined at all.}   Meanwhile, as a template for further size reductions, inertial contributions to rolling become less significant as size is reduced.  \hl{Small-scale interfacial forces are complex, but the authors' laboratory has previously measured adhesive forces on millimeter-scale micro-robot legs ranging from 0.04 to 1.5~mN/mm$^2$~\cite{Ryou}~\cite{Qu}; naive scaling of this two-mode robot's inertia and surface area downward would result in such interfacial forces matching inertial forces at a length scale of approximately 2 to 10~mm.  Approaching such scales, rolling locomotion would require alternate strategies for} leg coordination; more thorough incorporation of robot dynamics with controller behavior could be expected to increase capabilities.

\section{Conclusion}
We demonstrate walking/rolling motion for a 110 mm robot with just two degrees of freedom and finite joint range of motion.   The controller leverages geometry of non-uniform-radius robot legs to permit sustained walking or rolling locomotion plus rapid transitions between modes. Such closed-loop performance has not been demonstrated from multi-modal walking/rolling robots at comparable scales, and resulting costs-of-transport are competitive with other similarly-scaled robots for which results are available in literature. The robot's novel leg geometry and controller geometry may provide a template for further miniaturization of multi-modal mobile robots, by accommodating limited actuator capabilities and rudimentary sensing elements. 

\bibliographystyle{IEEEtran}
\bibliography{main}





\end{document}